# Smooth Like Butter: Evaluating Multi-Lattice Transitions in Property-Augmented Latent Spaces

Martha Baldwin,[1] Nicholas A. Meisel,[2] Christopher McComb[1]

[1]Department of Mechanical Engineering,
Carnegie Mellon University, Pittsburgh, PA 15213
[2]School of Engineering Design and Innovation,
The Pennsylvania State University, University Park, PA 16802

## Abstract

Additive manufacturing has revolutionized structural optimization by enhancing component strength and reducing material requirements. One approach used to achieve these improvements is the application of multi-lattice structures, where the macro-scale performance relies on the detailed design of mesostructural lattice elements. Many current approaches to designing such structures use data-driven design to generate multi-lattice transition regions, making use of machine learning models that are informed solely by the geometry of the mesostructures. However, it remains unclear if the integration of mechanical properties into the dataset used to train such machine learning models would be beneficial beyond using geometric data alone. To address this issue, this work implements and evaluates a hybrid geometry/property Variational Autoencoder (VAE) for generating multi-lattice transition regions. In our study, we found that hybrid VAEs demonstrate enhanced performance in maintaining stiffness continuity through transition regions, indicating their suitability for design tasks requiring smooth mechanical properties.

## 1. Introduction

Additive manufacturing (AM) has enabled more design freedom for engineers; however, these freedoms have required designers to become more creative when developing solutions due to continued pushes for optimization and material reduction. One approach to keep pace with these advancements is to utilize lattice structures, which can be used to reduce weight while maintaining performance.[1,2] By simply applying lattice structures to a design, it is possible to decrease weight by as much as 40% while maintaining overall strength.[3] This is often accomplished using *uniform lattice* structures, which are composed of a single type of unit cell with the same density and size throughout the structure.[4] However, these structures are limited by the properties of a single unique lattice[1,5,6], which places some restrictions on the design space. This work explores machine learning models to enable the facile design of multi-lattice structures, with a long-term goal of supporting designers in taking advantage of the freedom enabled by AM.

### 1.1 Motivation for Multi-lattice Structures

*Graded lattice* structures enable a wider range of mechanical properties than uniform lattice structures alone.[2,4,7–11] *Graded lattice* structures are composed of a single unit cell topology but with a varying volume fraction or relative density.[8,10,12] Due to the wide range of properties achievable through graded lattice structures, they can be designed to exhibit higher stiffness than uniform lattice structures.[4,7,9,10,12] Additionally, these structures have been cited as promising for impact reduction applications[2,11,13] based on their ability to absorb more energy than uniform lattice structures[7,11], exceeding the compressive qualities of uniform lattices by up to 25%.[14] These traits suggest that graded lattice structures could be useful in cases of dynamic loading, such as surgical implants[15] and impact protection equipment.[2,11,13]

However, there are two major limitations to implementing graded lattice structures: they are difficult to design, and their performance is directly dependent on the type of unit cell chosen. The difficulty surrounding design is due to the complexity of these structures in conjunction with a lack of available DfAM software that can handle these complexities. Although these structures may initially appear to be straightforward variations of uniform lattices, it can still be difficult to ensure that they are ultimately manufacturable, which is evident based on the number of works dedicated to improving the connectivity of graded lattice structures.[14,16,17] Even when these design limitations can be overcome, graded lattice structures are still restricted by their dependence on the chosen unit cell topology. For instance, Panesar et al. found that surface-based unit cells have better connectivity than strut-based unit cells.[4] As an example of another constraint, Plocher et al. noted that the degree of gradation of the structures did not affect Schwarz-P lattices, but did directly influenced the cumulative energy absorption of body-centered cubic lattices.[2] A similar work emphasized this point by demonstrating that various lattice types of the same density gradient exhibited different performance in terms of energy absorption.[7] The accumulation of these constraints restricts designers to a single type of microstructure, which ultimately reduces performance.[18,19] These cases motivate the exploration of more variable and complex lattice structures.

The overall motivation for researching graded lattice structures has been the increased design freedom combined with the many benefits they have over uniform lattice structures.[2,4,7–12] However their limitations have encouraged designers to explore other alternatives, such as developing structures with multiple types of unit cell topologies[13,15,18–23] which are often referred to as *multi-lattice* structures. A *multi-lattice* structure is a lattice structure that is composed of multiple types of unit cell topologies.[15,20,23] Multi-lattice design has become a major area of interest in AM due to its versatility in terms of properties and mechanical behavior. Some work has shown that random lattice structures are better suited for energy absorption applications[24] which aligns well with works that recommend multi-lattice structures for impact reduction applications.[13] Most of the research that implements multi-lattice structuring is restricted to unit cells that have inherently overlapping boundaries to ensure effective connectivity between adjacent unit cells.[13,15,18–21,25,26] The first works to validate the use of such structures utilized compatible boundaries in a 2.5-dimensional space to demonstrate that multi-lattice structures could outperform uniform lattice structures in terms of stiffness and strength[13,20,21], as well as maximum stress and weight.[15] There have also been 3-dimensional cases using compatible boundaries that proved that the multi-lattice structures exhibited improvements over uniform lattice structures in terms of strength[25], elastic modulus, and energy absorption.[23]

### 1.2 Designing Multi-lattice Structures

Although multi-lattice structures have significant adaptability in geometry and mechanical properties, they are still extremely difficult to design properly. If the boundaries between adjacent unit cells do not align properly it can lead to stress concentrations that negatively impact the overall strength of the structure.[20] Wang et al. have described the development of smooth transition regions as a multi-scale design problem which must be addressed at the micro and macro-scale.[27] Therefore, the majority of methods used for creating multi-lattice structures use density-based topology optimization to assign a set of unit cells to a structure.[13,20,25,26,28] However, the performance of this approach is dependent on the selection of a compatible and optimal set of unit cells. The common approaches for choosing this set of unit cells consist of numerical[13,19,22,28–32] and machine learning approaches.[21,27,33–37] In order to compare them, we propose three criteria to evaluate multi-lattice structures to determine if they are suitable:

① maintain connectivity between adjacent unit cells[20];
② consist of a wide range of unit cell topologies; and
③ are designed based on the mechanical characteristics of adjacent unit cells.[27]

These design criteria provide a common point of comparison for all the models discussed in the current work. Few numerical approaches address a combination of the criteria well. For instance, a parameterized lattice design approach was successfully implemented in several works, where the lattices were designed based on a set of parameterizable characteristics.[13,19,29] However, these approaches require that the set of lattices be parameterizable, which is not possible for all lattice types, therefore this approach only meets criteria ③. Sanders et al. designed a numerical approach to creating interpolations between unit cells using signed distance functions, however it is limited to unit cells composed of struts, bars or plates.[22] Unfortunately, this approach is only suitable for creating structures that meet design criteria ①. An approach by Chan et al. utilizes shape blending to create smooth transitions between adjacent unit cells.[28] This demonstrated a working numerical design method based on geometry for developing transition regions that addressed design criteria ②, but it does not guarantee the other design criteria. Several other numerical approaches are capable of addressing design criteria ①[30] and ③ concurrently using shape blending approaches, but only for TPMS lattice structures.[31,32] Additionally, these shape blending approaches result in potentially asymmetric lattice structures which nullify the cubic assumption, thus increasing the complexity of the stiffness tensor for finite element analysis.[32] This motivates the use of approaches that create transition regions using a progressive series of lattices, which may be executed most effectively by data-driven methods.

Due to the complex nature of this design problem, researchers have explored approaches that use machine learning or data-driven methods to design multi-lattice transition regions.[27,35–37] Most data-driven approaches are centered around reducing the dimensionality of the data in order to diminish the computational costs of organizing sets of unit cells for multi-lattice structures.[21,27,35–37] There are also a few prominent machine learning models that have proven successful, including variational autoencoders (VAEs)[27] and generative adversarial networks (GANs).[35] One of the first approaches using data-driven design creates a reduced representation of the unit cells using a Laplace-Beltrami spectrum, which is combined with a neural network to develop new unit cells.[36] During the development of structures, they are limited to classes of unit cells in order to guarantee connectivity. However, this work does successfully address design criteria ① and ③. Another technique was to use the latent-variable gaussian process to develop a latent space for generating sets of unit cells.[21,37] These approaches enforce property continuity through the topology optimization step for the structure and use strut-based datasets that are compatible with the latent-variable gaussian process.[21,37] Therefore, these models are only able to address criteria ③. A model using an inverse homogenization generative adversarial network (IH-GAN) generates a set of unit cells that are compatible with respect to their mechanical properties.[35] The model represents the unit cells in a reduced form by using unit cells that have parameterizable characteristics. Therefore this model is best designed to address criteria ③, but it is restricted to parameterizable unit cells and does not guarantee connectivity of consecutive unit cells without post-processing.

It currently seems that only one data-driven approach, variational autoencoders (VAEs), can address design criteria ② when it comes to selecting unit cells for density-based topology optimized structure design.[27,33,34] The first implementation of a VAE for this task was executed by Wang et al., which found that classes of compatible unit cells could be generated by traveling through the latent space generated by the VAE.[27] Although various classes can be created for any type of unit cell, it restricts the ability to generate structures containing two drastically different topologies. The development of these classes was done to address the connectivity of unit cells, while also enabling a wide variety of mechanical properties among the unit cell set. Therefore, this approach directly addresses design criteria ③, and partially addresses design criteria ② and ①. Given that VAEs demonstrate the most promise for developing a diverse, connected, and functional set of unit cells, it is important to fully understand what affects their performance.

### 1.3 Research Questions

The overarching goal of this work is to improve the quality of a set of lattices chosen for use with a density-based topology optimization algorithm, evaluated on design criteria ① and ③. We propose to accomplish this goal via a method that uses VAEs to address these design criteria concurrently. This work will propose a method for evaluating the continuity of mechanical performance of a set of unit cells (e.g., design criteria ③). In our prior work, we developed a method for measuring the similarity of consecutive cells, which serves as an approach for partially measuring criteria ①.[33,34]

This study specifically aims to determine the benefits of incorporating both mechanical properties and geometrical data into VAEs. Understanding how to properly navigate the latent space is crucial for developing functionally ③ and geometrically smooth ① multi-lattice transition regions. Prior work conducted using a strut-based dataset and a geometry-driven VAE found that the performance of geometric smoothness was primarily attributed to the latent space distance between two unit cells in the latent space and partially affected by the length of the transition region.[33,34] The current work significantly builds on that prior work by examining how these relationships change if the latent space is informed by both geometry and property data. Specifically, we examine *stiffness continuity*, a measure of the change in stiffness across adjacent unit cells, as an indicator of the stiffness throughout a transition region. Our primary research questions address the unknowns regarding the role of mechanical properties in lattice cell latent spaces:

1. How does incorporating mechanical properties into VAEs affect the relationship between *geometric smoothness*, distance, and transition length in the latent space?
2. How does incorporating mechanical properties into VAEs affect the relationship between *stiffness continuity*, distance, and transition length in the latent space?

### 2. Methodology

This section outlines the architectures developed and the methods of testing used to evaluate the performance of the transition regions of various VAEs. Additionally, it will briefly outline our use and development of data for this work. The methods discussed in this work are summarized in Figure 1.

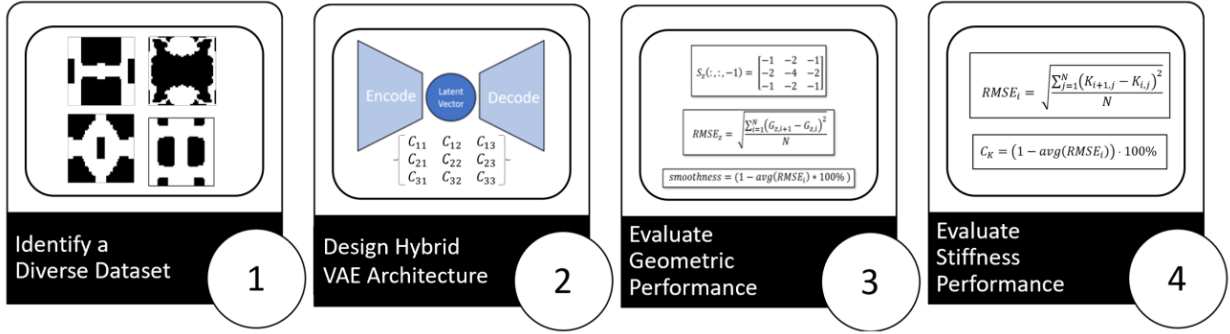

**Figure 1:** Illustration of Methodologies

## 2.1 Unit Cell Dataset

The data used here was originally generated in prior work by Wang et al. and consists of 248,396 unique unit cells and their corresponding stiffness tensors.[27,36,38] Each unit cell is represented as a binary array of size 50×50, where a value of 0 represents material absence and a value of 1 represents material presence. Figure 2 illustrates the diversity of the unit cells in this dataset. Additionally, each unit cell is associated with a corresponding stiffness tensor, which is represented as a 3×3 array of floating-point values.

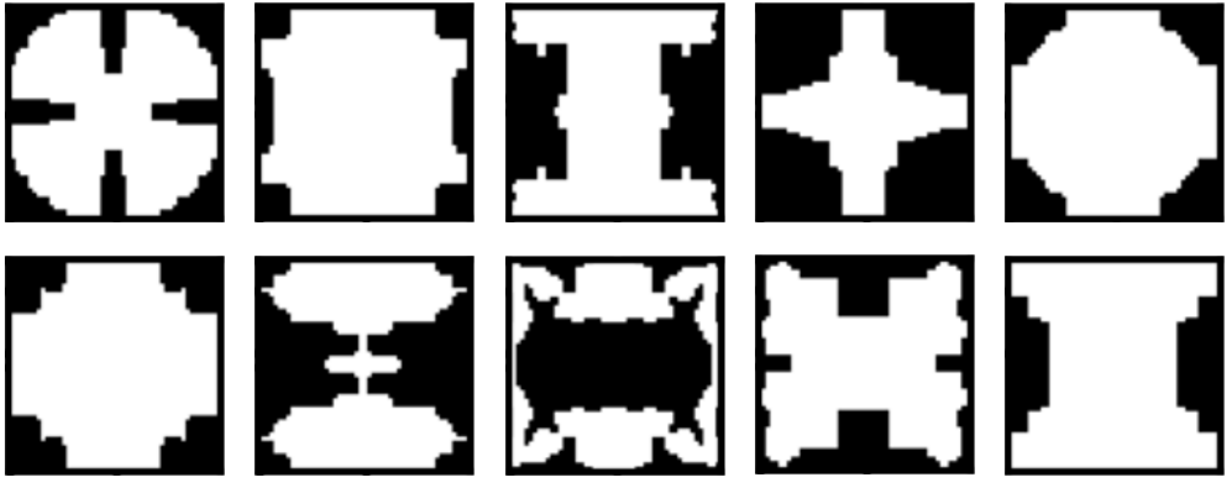

**Figure 2:** Sample of unit cells from dataset by Wang et al.[27,36,38]

## 2.2 Architectures

This work employs a variational autoencoder (VAE) for developing multi-lattice transition regions. VAEs are machine learning models that perform a data-driven dimensionality reduction in order to make inferences more efficiently.[39] The dimensionality reduction is executed by an encoder that creates a latent representation of the data.[40,41] The data is then reconstructed from the latent representation by training a decoder based on the error between the reconstruction and the true image. The VAE ultimately creates a generative latent space which can be used to generate the transition region between two unit cells.

This work specifically utilized two neural network architectures: a geometry-only VAE based on prior work, and a hybrid representation VAE that encodes both stiffness and geometry information. Both architectures were trained using identical hyperparameters: a batch size of 32, the Adam optimizer[42], and an 85%/15% training/testing split of the dataset. Additionally, training was terminated early if the loss failed to improve after 10 full epochs, where the loss term measures the difference between the reconstructed data and the original data. Both models were further instantiated as β-VAEs, a type of VAE in which a constant value, β, is used to weight the KL-divergence loss of the model.[43] We specifically used a value of $\beta_{NORM} = \beta * \cdot D/W$, where $D$ is the latent space dimensionality and $W$ is the image width, which is recommended by Higgins et al.[43] Finally, this work uses a 16-dimensional latent representation based on prior works using it for reducing dimensionality of the same[27] or similar datasets.[36]

The geometry VAE simply encodes and decodes the geometry (see Figure 3). The architecture of the model was based on our previous works[33,34], which provide more detailed outlines of the architecture. The purpose of using the model in this work is to serve as a baseline for comparison against the hybrid architecture.

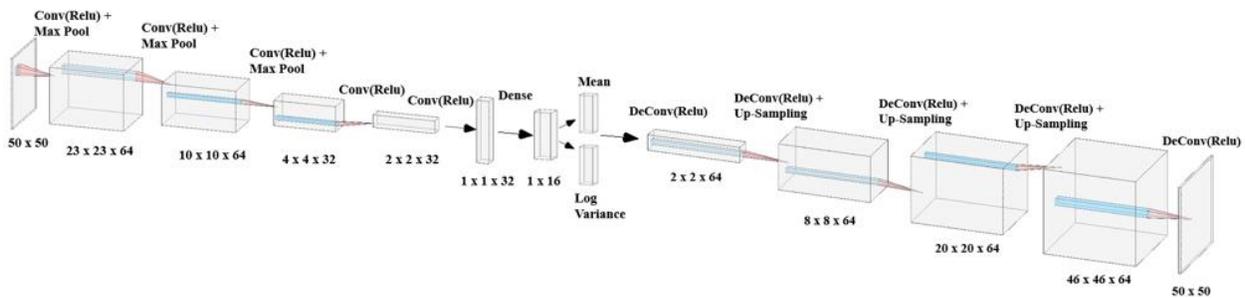

**Figure 3:** Geometry VAE Architecture* where the input consists of only the unit cell geometry

The hybrid VAE (see Figure 4) has a similar encoder and decoder framework to the geometry VAE. However, the encoded geometry is appended with a vector containing additional information about the corresponding stiffness of the input geometry. The stiffness tensor is first flattened and then fed through a linear layer which allows this information to be better integrated into the resulting latent space. The decoder of this architecture is identical to the decoder of the geometry decoder architecture to enable consistent comparison.

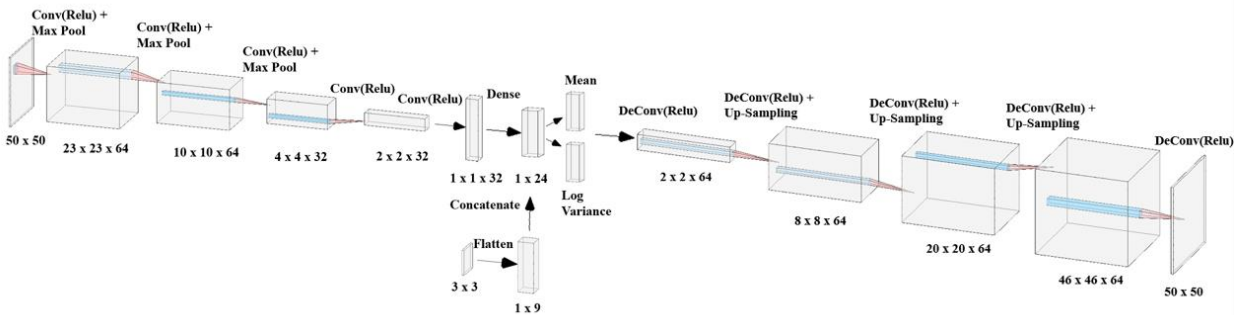

---

\* Image created with http://alexlenail.me/NN-SVG/AlexNet.html

**Figure 4:** Hybrid VAE Architecture* where the input consists of unit cell geometry and the corresponding stiffness tensor

Through the training process, the VAE generates a latent space where interpolations can be developed to create transition regions between lattices. The desired endpoints of the transition region are converted to latent points by the encoder. The latent points of the intermittent unit cells are then determined through a linear interpolation between the latent endpoints, performed in latent space. The decoder is then used to reconstruct the unit cells of the corresponding latent points, producing a complete transition region.

### 2.3 Evaluating Performance of Architecture

The interpolations generated with the geometry VAE and hybrid VAE are evaluated in terms of both geometric smoothness, design criteria ①, and stiffness continuity, design criteria ③. Prior work establishes a procedure for evaluating geometric smoothness and proved that it is related to the latent space distance and transition length.[33,34] We use the same analysis procedure and geometry smoothness metric here. Specifically, latent space distance is measured with respect to number of standard deviations in the latent space, and transition length is simply the number of points in a transition region. We also introduce a new metric for evaluating stiffness continuity in order to determine the relationship between stiffness continuity, latent space distance, and transition length.

#### 2.3.1 Evaluating Geometric Smoothness

The purpose of this metric is to evaluate the similarity of consecutive unit cells in the transition region, which partially measures connectivity of adjacent unit cells, design criteria ①. The success of a latent space is dependent on the performance of the interpolations that can be produced from that latent space. When evaluating geometric transitions in a latent space, continuously and smoothly changing geometries are desired. To perform this evaluation, we utilize a smoothness metric that was developed in our previous work to evaluate the geometric smoothness of a 2D interpolation.[33,34] This metric calculates the gradients between multiple layers in an interpolation, essentially measuring the flow between each layer of pixels. The root mean squared error (RMSE) is calculated between the flows (Eq (1)) and then normalized (Eq( 2)) to produce a value of smoothness over an entire interpolation (4). This metric will be utilized in this work to evaluate the geometric smoothness of interpolations among the various models, which measures the similarity of consecutive unit cells. More details on the implementation of this smoothness evaluation are available in prior work by the authors.[33]

$$RMSE_{x,i} = \sqrt{\frac{\sum_{j=1}^{N}(G_{x,i+1,j} - G_{x,i,j})^2}{N}}$$

$$RMSE_{y,i} = \sqrt{\frac{\sum_{j=1}^{N}(G_{y,i+1,j} - G_{y,i,j})^2}{N}} \quad (1)$$

$$RMSE_{z,i} = \sqrt{\frac{\sum_{j=1}^{N}(G_{z,i+1,j} - G_{z,i,j})^2}{N}}$$

where $RMSE_x, RMSE_y,$ and $RMSE_z$ are the root mean squared errors of a pair of gradients in the $x$, $y$, and $z$ directions respectively, $G_x, G_y, and\ G_z$ are the gradient array components in the $x$, $y$, and $z$ directions respectively, $i$ is the index between each gradient array and their respective images, $j$ is the index that identifies the specific term in the gradient array, and N is the number of terms in a single gradient array. The normalization of the RMSE is required for the smoothness value to be represented as a percentage.

$$RMSE_i = \frac{RMSE_{x,i} + RMSE_{y,i} + RMSE_{z,i}}{3 \cdot RMSE_{max}} \qquad (2)$$

where $RMSE_{max}$ is the maximum possible $RMSE_i$ which is calculated based on the filter used. $RMSE_i$ is then averaged to compute the average over the entire transition region.

$$\overline{RMSE} = \frac{1}{n-3}\sum_{i=1}^{n-3} RMSE_i \qquad (3)$$

where $\overline{RMSE}$ is the average RMSE over the entire transition region and $n$ is the total number of unit cells in a transition region including endpoints.

$$C_S = (1 - \overline{RMSE}) \times 100\% \qquad (4)$$

where $C_S$ is a value representing the smoothness of geometry over the entire transition region. Higher values indicate a smoother and more continuous transition, while lower values indicate abrupt transitions.

### 2.3.2 Evaluating Stiffness Continuity

The purpose of this metric is to measure the ability of the model to produce transitions that promote smooth mechanical characteristics, design criteria ③. In contrast to the geometric smoothness metric, this metric directly computes an evaluation for design criteria ③. This evaluation is performed on the calculated stiffness tensors of the reconstructed data, in order to have accurate stiffness tensors. When evaluating the equivalent of geometric smoothness for stiffness tensors, the changes within a single tensor are not important as each element represents a unique piece of information. Therefore, the tensors are evaluated by comparing values only with the neighboring tensor. To achieve this, we implemented an RMSE framework to calculate a stiffness continuity value. First, the stiffness tensors in the interpolation were normalized with respect to the entirety of the training data. Then, the RMSE must be computed between each pair of values in the stiffness tensor using:

$$RMSE_i = \sqrt{\frac{\sum_{j=1}^{N}(K_{i+1,j} - K_{i,j})^2}{N}} \qquad (5)$$

where $RMSE_i$ denotes the root mean squared error between a pair of stiffness tensors of unit cells $i$ and $i+1$, $K$ is the normalized flattened stiffness tensor, $j$ is the index that identifies the specific term in flattened stiffness tensor, and $N$ is the number of terms in a single stiffness tensor (in this case, 9).

Since the stiffness tensors are normalized before computing the RMSE, the maximum possible value is 1. This allows the direct evaluation of the continuity of the stiffness as a percentage by averaging the values using Equation 3:

$$C_K = (1 - \overline{RMSE}) \times 100\% \qquad (6)$$

where $C_K$ is a value representing the continuity of stiffness over the entire transition region. Higher values indicate a smoother and more continuous transition in stiffness, while lower values indicate abrupt transitions. The interpretation of this metric aligns with that of geometric smoothness.

## 3. Results

In this section we discuss the evaluation of these models in four main areas: the reconstruction capabilities, the test interpolations described in the methods, an ordinary least squares evaluation based on the interpolation results, and sample transition regions. The ordinary least squares regression analysis is used to determine the ability of the model to address the design criteria: ① maintain connectivity between adjacent unit cells[20] and ③ designed based on the mechanical characteristics of adjacent unit cells.[27]

### 3.1 Model Reconstruction

Initially, we evaluate the performance of a machine learning model by examining the performance during training and testing, here using a mean squared error (MSE) and KL-divergence loss, and coefficient of determination. This indicates the ability of the model to appropriately reconstruct the desired data. If the model can reconstruct data, then an examination of the learned latent space may be informative. The results for the Geometry VAE are shown in Figure 5. This model serves as the baseline and aligns with the model originally developed in previous work.[33,34] Specifically, these results indicated that the unit cell geometry can be effectively reconstructed to 95% accuracy for both testing data and training data.

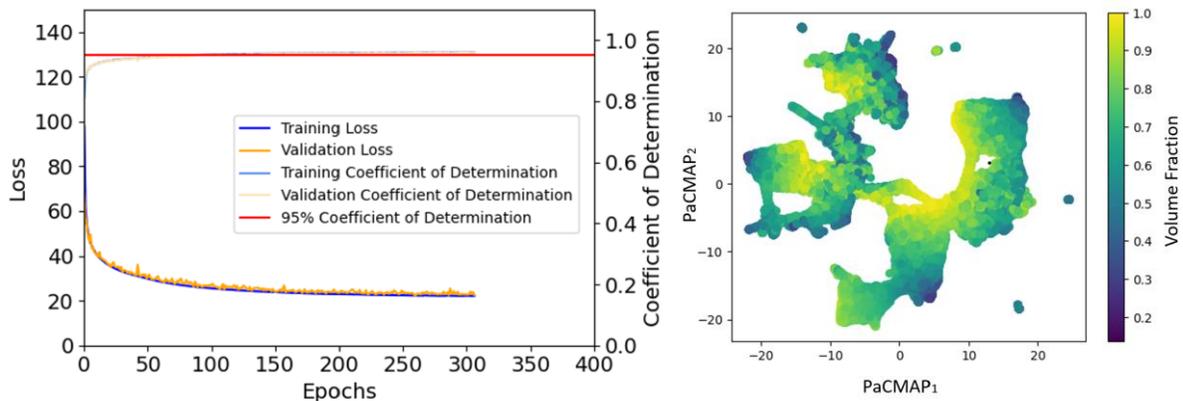

**Figure 5:** Geometry VAE: Plot of Loss and Coefficient of Determination (left) and Visualization of the Latent Space using PaCMAP Dimensionality Reduction (right)

The performance of the hybrid VAE was highly similar to the geometry VAE, with matching accuracy for both training and testing data (see Figure 6). This is a good indication that the model will perform well when reconstructing data, and the results in the evaluation section should be directly comparable. However, the hybrid VAE did reach the stopping criteria earlier.

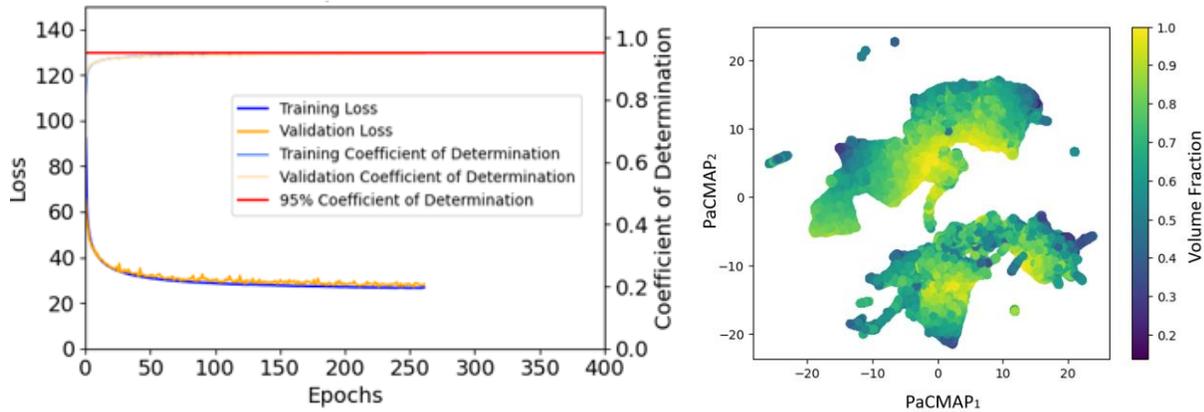

**Figure 6:** Hybrid VAE: Plot of Loss and Coefficient of Determination (left) and Visualization of the Latent Space using PaCMAP Dimensionality Reduction (right)

In addition, a Pairwise Controlled Manifold Approximation (PaCMAP) was used to visualize the latent space produced by each of the trained models (see the right images of Figure 5 and Figure 6).[44] The axes in these figures are the first and second dimensions of the PaCMAP embedding, respectively. This visualization approach has been shown in prior work to accurately provide a representation with balanced preservation of local and global features for engineering-relevant data.[45] There are distinct similarities in the learned embeddings, despite the addition of unique mechanical performance information in the hybrid model. The overall shape is similar, consisting of four distinct clusters. In addition, many of the unit cells with lower volume fraction are placed at the periphery of the latent space.

### 3.2 Interpolation Performance

In order to evaluate the interpolation performance, evenly-spaced interpolations were produced using an increasing number of standard deviations to provide a standardized distance metric. This ensures that the analysis is repeatable and comparable among multiple latent spaces. The interpolations are generated relative to the mean of the latent space, originating at -3 standard deviations and iteratively extended up to +3 standard deviations. Each interpolation is then produced using 5, 10, and 15 unit cells.

Table 1 and Table 2 show a set of interpolations with a length of 10 unit cells. In the images that make up these tables, black pixels represent material absence and white pixels represent material presence. The first and last images in the tables represent the endpoints of the transition region. The full set of transitions are evaluated and plotted in Figure 7 and Figure 8 respectively.[30,31] Each point in these plots represents a single linear interpolation, with latent space distance shown on the x-axis and the average value of the corresponding metric (either geometric smoothness or stiffness continuity) shown on the y-axis. The points are labeled using color to denote the length of the transition region.

**Table 1:** Geometry VAE: Visualization of interpolations based on number of standard deviations with 10 transition intervals

| Number of Standard Deviations Between Endpoints | Resulting Interpolations | Geometric Smoothness, $C_S$ (%) | Stiffness Continuity, $C_K$ (%) |
|---|---|---|---|
| 1 | 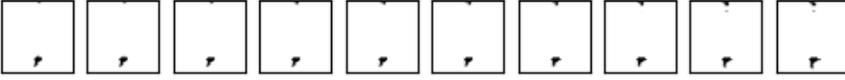 | 98.11 | 99.58 |
| 2 | 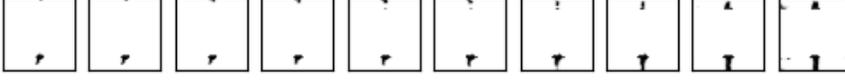 | 95.12 | 99.06 |
| 3 | 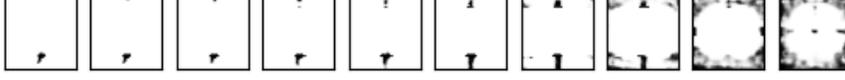 | 90.07 | 97.20 |
| 4 | 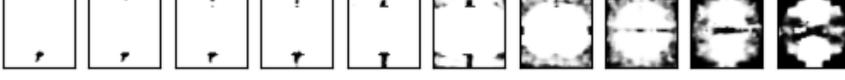 | 84.48 | 94.57 |
| 5 | 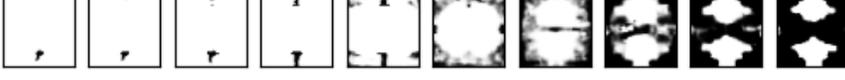 | 80.70 | 94.20 |
| 6 | 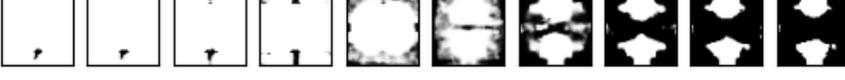 | 78.35 | 94.25 |

**Table 2**: Hybrid VAE: Visualization of interpolations based on number of standard deviations with 10 transition intervals

| Number of Standard Deviations Between Endpoints | Resulting Interpolations | Geometric Smoothness, $C_s$ (%) | Stiffness Continuity, $C_K$ (%) |
|---|---|---|---|
| 1 | 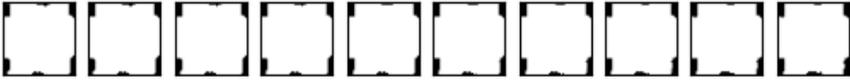 | 98.08 | 99.92 |
| 2 | 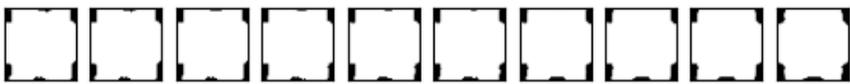 | 96.27 | 99.75 |
| 3 | 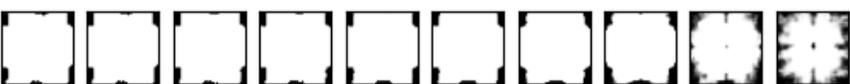 | 91.70 | 99.03 |
| 4 | 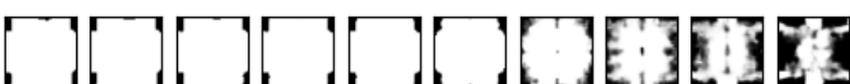 | 85.85 | 97.25 |
| 5 | 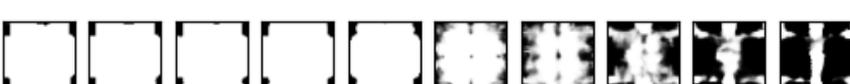 | 80.43 | 96.58 |
| 6 | 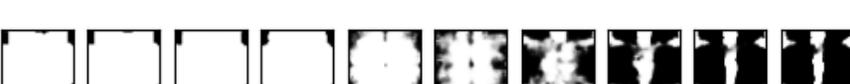 | 77.18 | 96.40 |

The transition regions were first evaluated using the geometric smoothness metric outlined in the methods section, this metric is discussed in further detail in prior work.[33,34] The results from Figure 7 (left) are consistent with the results seen in previous work.[33,34] As latent space distance increases, the smoothness decreases. Additionally, the higher transition lengths have higher smoothness values relative to one another. The predicted geometries from Figure 7 (left) were used to the calculate the corresponding stiffness tensors, which were evaluated using the stiffness continuity metric described in the methods section to produce Figure 7 (right). This figure serves as the baseline of stiffness continuity for our hybrid geometry/property model. From visual observations, the stiffness continuity begins to plateau as the latent space distance increases. This indicates that the effect of latent space distance is limited to some degree.

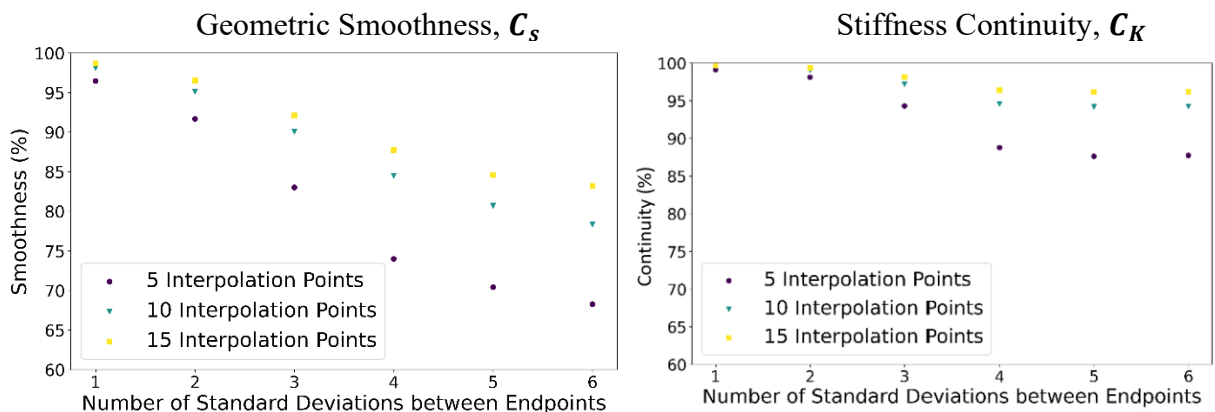

**Figure 7:** Geometry VAE: Geometric Smoothness versus Number of Standard Deviations in the Latent Space (left) and Stiffness Continuity ($C_K$) vs Number of Standard Deviations in the Latent Space (right).

The results from the hybrid VAE are consistent with the results from the geometry VAE, where the relationship between smoothness, latent space distance, and transition length are comparable (see Figure 8 (left)). This is a desirable trait, as it indicates that the latent space has similar embeddings to the baseline model. This means that we can directly compare the performance of the two models in terms of the stiffness continuity. The predicted geometries from Figure 8 (left) were used the calculate the corresponding stiffness tensors, which were evaluated using the metrics described in the methods section to produce Figure 8 (right). The stiffness continuity of the data in Figure 8 (right) appears to plateau much sooner than the points in Figure 7 (right).

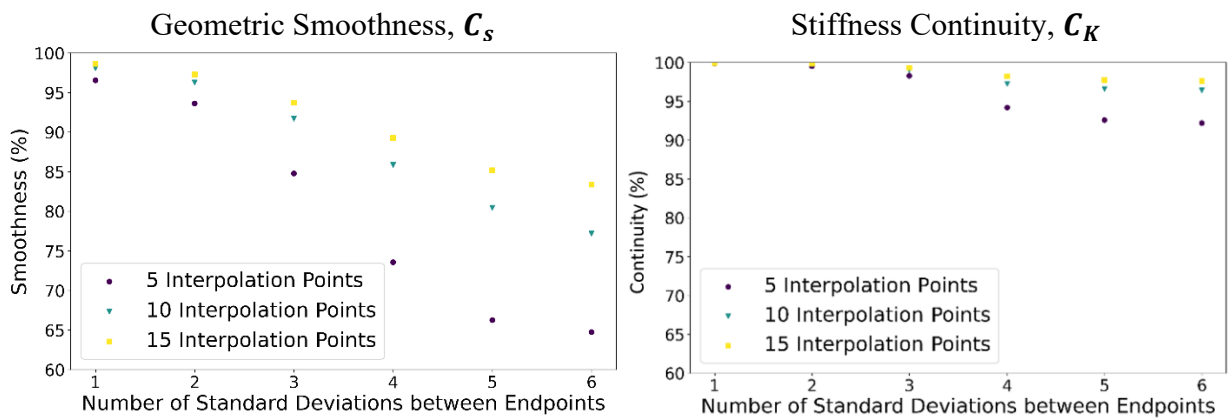

**Figure 8:** Hybrid VAE: Geometric Smoothness versus Number of Standard Deviations in the Latent Space (left) and Stiffness Continuity ($C_K$) vs Number of Standard Deviations in the Latent Space (right).

Visually, Figure 7 (left) and Figure 8 (left) are consistent with prior work, showing that geometric smoothness has a clear relationship to latent space distance and the length of the transition region. Geometric smoothness appears to be positively correlated with the length of the transition region, and negatively correlated with the distance in the latent space between the endpoints. Figure 7 (right) and Figure 8 (right) show the relationship between stiffness continuity, latent space distance, and length of the transition region. These plots demonstrate that the effect of latent space distance and transition length does not have as significant of an impact as they do on geometric smoothness. However, these values may be somewhat inflated given that symmetric geometries contain zeros in the stiffness tensor.[46] By visually comparing the two plots, it appears as though the hybrid model performs better, as it plateaus sooner. The next section serves to further understand the capability of these models and Section 3.4 explores the validity of these visual observations.

### 3.3 Sample Transition Regions

This section examines a variety of sample interpolations to compare the qualitative performance of the geometry and hybrid VAE models. These samples serve to demonstrate the capability of these models for developing sets of unit cells for density-based topology optimization, but also for developing smoothly varying space-filling structures.

Specifically, in Table 3 we depict transition regions for which both endpoints are in the same latent space cluster (referred to as *intra*-cluster) and those for which endpoints are in separate clusters (referred to as *inter*-cluster transitions). The clusters that we refer to are groupings observed in the latent space in Figure 4 and Figure 5. The predicted geometries themselves and the geometric smoothness of the corresponding transition regions are all comparable. Inter-cluster transitions are somewhat less consistent. We attribute this to the potential for poor reconstructions when interpolating between clusters, as there is not sufficient training data in those regions.

A mesh interpolation was performed within and between clusters to further compare the performance of the hybrid and geometry VAE embeddings (see Figure 9). A mesh structure can be generated by producing an interpolation between four corner unit cells, resulting in an interpolation region with both width and height. Visual analysis of the intra-cluster structures, shown in Figure 9(a) and Figure 9(b), reveals few differences between the models, where each unit cell is rather comparable between the two figures. The similarities between the structures suggest that the latent embeddings in both models were extremely similar. The inter-cluster structures, shown in Figure 9(c) and Figure 9(d), reveal that some unit cells are significantly distorted. This is likely due to the lack of training data between clusters. Overall, the mesh interpolations in Figure 9 illustrate the ability to produce entire structures, while outlining the potential issues with manufacturing using this approach.

Table 3: Example Transitions

| Type of Model: | Relation between Clusters (Standard Deviations between Points): | Example Transitions (Geometric Smoothness, $C_s$): |
|---|---|---|
| Geometry VAE | Intra-Cluster Interpolation (4.48 std) | 86.10% 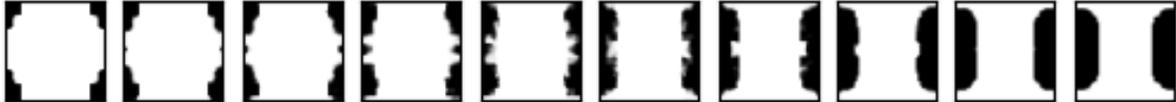 |
| Geometry VAE | Inter-Cluster Interpolation (6.55 std) | 83.78% 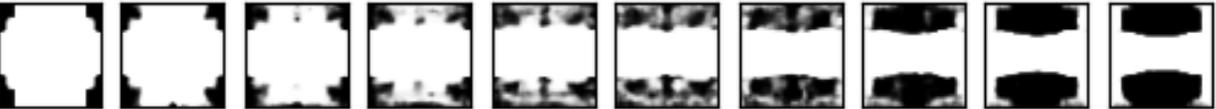 |
| Hybrid VAE | Intra-Cluster Interpolation (4.58 std) | 86.57% 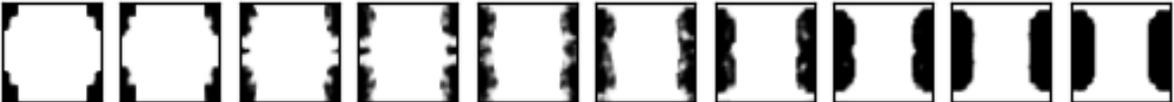 |
| Hybrid VAE | Inter-Cluster Interpolation (5.95 std) | 81.96% 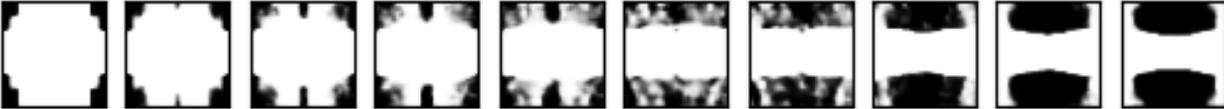 |

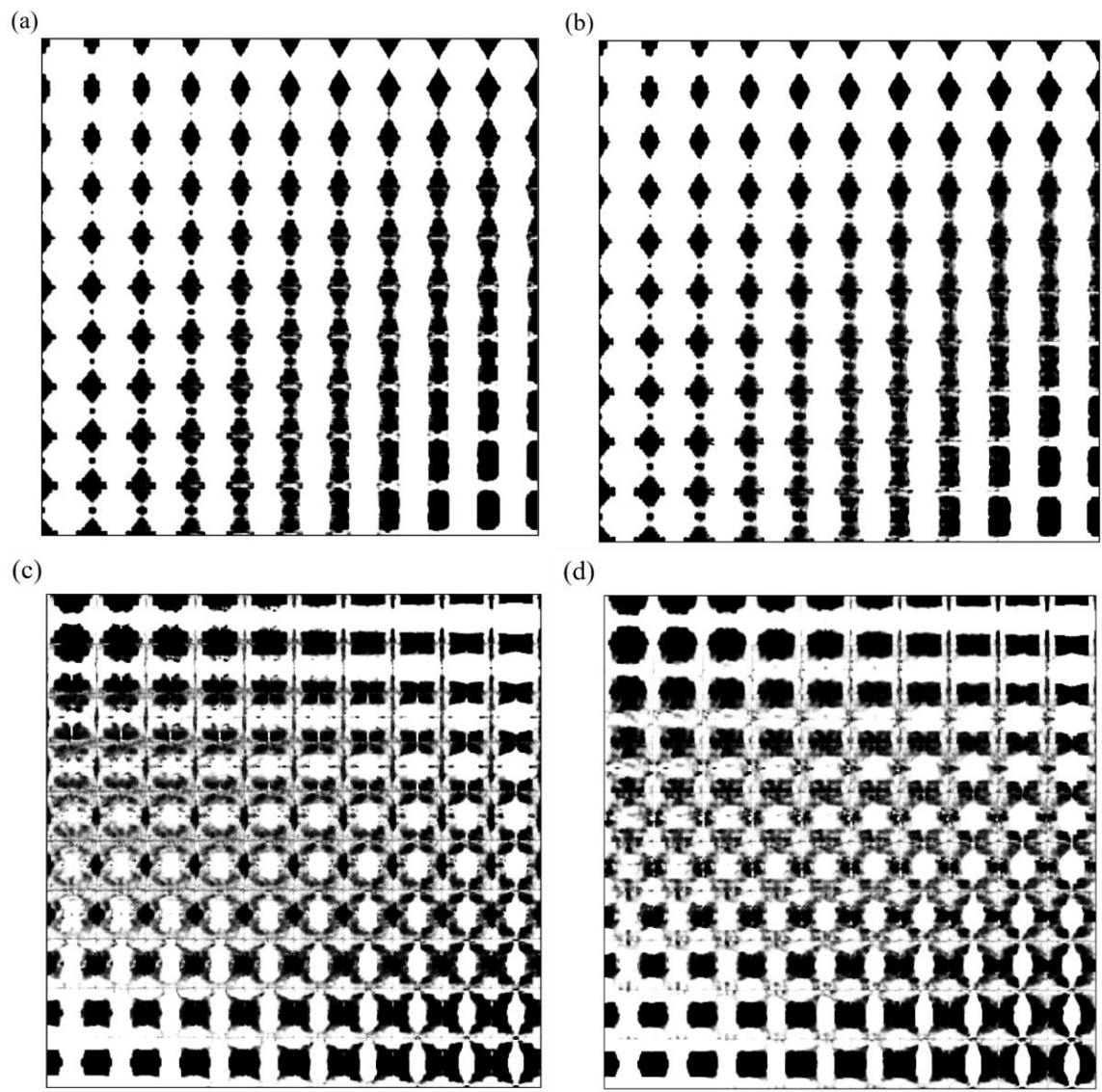

Figure 9: Examples of Mesh Interpolation- (a) Geometry VAE Embedding within Cluster (b) Hybrid VAE Embedding within Cluster (c) Geometry VAE Embedding between Clusters (d) Hybrid VAE Embedding between Clusters

### 3.4 Evaluating Latent Space Relationships

The smoothness of the geometry throughout a transition region partially describes the connectivity of unit cells, design criteria ①, and the continuity of stiffness is representative of the ability to choose unit cells based on mechanical properties, design criteria ③. Therefore, we conduct a series of linear regression analyses using those attributes as the dependent variables. The independent variables are latent space distance and the length of the transition region. Table 4 displays the ordinary least squares regression analysis of the data points in Figure 7 and Figure 8, which represent the performance of the geometry VAE and hybrid VAE respectively. Highly significant terms are denoted by *, which represent a p-value less than 0.05.

The results of the geometry VAE serves as the baseline since its architecture is most similar to that used in prior work.[33,34] The first column in Table 4 displays the ordinary least squares regression analysis for the relationship between geometric smoothness and latent properties for the geometry model. The second column in Table 4 displays the ordinary least squares regression analysis for the relationship between stiffness continuity and latent properties for the geometry model. The results from first column in Table 4, are consistent with prior work, where the transition length alone does not have a significant relationship with geometric smoothness and instead latent space distance has the most significant impact on geometric smoothness.[33,34] This is evident based on the p-values in first column in Table 4. The high R-squared value indicates that the variability in the geometric smoothness is adequately described by the latent space distance and transition length. Based on the results of first and second columns in Table 4, the relationship between stiffness continuity and the properties of the geometry defined latent space are the same as the relationship with geometric smoothness. This is based on the high R-squared value of the stiffness continuity of the geometry VAE in the second column in Table 4 and the corresponding p-values. The coefficients in the first and second columns in Table 4 suggest that latent space distance has less impact on the stiffness continuity than it does on geometric smoothness, which confirms the visual observations made on Figure 7. An R-squared value of 91% indicates that 9% of the variability of the stiffness continuity is not accounted for by the two independent variables explored (see second column in Table 4). This suggests that there are additional variables that may be affecting the stiffness continuity of the transition region.

The results of the hybrid VAE are displayed in the third and fourth columns of Table 4. The third column in Table 4 displays the ordinary least squares regression analysis of the results from Figure 8 (left), which evaluates the relationships between geometric smoothness and latent properties for the hybrid model. The fourth column in Table 4 displays the ordinary least squares regression analysis of the results from Figure 8 (right), which evaluates the relationships between stiffness continuity and latent properties for the hybrid model.

**Table 4:** Ordinary Least Squares Regression Analysis for *Geometric Smoothness* and *Stiffness Continuity* versus Latent Space Distance and Transition Length

|  | Geometry VAE | | Hybrid VAE | |
| --- | --- | --- | --- | --- |
|  | Geometric Smoothness | Stiffness Continuity | Geometric Smoothness | Stiffness Continuity |
| R-Squared: | 0.961 | 0.914 | 0.957 | 0.908 |
| Constant: | 102.0397±2.915* | 102.5917±1.989* | 106.2413±3.434* | 103.1278±1.171* |
| Number of Standard Deviations (Distance): | -7.3090±0.749* | -3.4465±0.511* | -8.8864±0.882* | -2.3195±0.301* |
| Transition Length: | 0.0225±0.270 | -0.1444±0.184 | -0.2198±0.318 | -0.1751±0.108 |
| Interaction Term: | 0.2747±0.069* | 0.1858±0.047* | 0.3864±0.082* | 0.1263±0.028* |

Highly significant terms are denoted by *, which represent a p-value less than 0.05.

The results illustrate that incorporating mechanical properties into the model did not change the relationships between geometric smoothness, latent space distance, and transition length. By comparing the p-values in Table 4, it is clear that both the geometry VAE and hybrid VAE are strongly affected by the latent space distance and the combination of latent space distance and transition length, but not transition length alone. Therefore, the hybrid and geometry VAE models perform similarly in terms of geometric smoothness. Interestingly, the stiffness continuity in the hybrid model has the same relationships based on the p-values from the fourth column in Table 4. By comparing the coefficients of latent space distance for the stiffness continuities, in the second and fourth columns of Table 4, there is a key difference in performance of the two models. The stiffness continuity is less negatively affected by latent space distance in the hybrid embedding. Even though the interaction term does offset some of these effects, the stiffness continuity still performs better in the hybrid model, which was evident in Figure 8 (right). However, the R-squared value of 90% suggest that other variables account for approximately 10% of the variability of stiffness continuity (see fourth column Table 4).

Overall the results indicate that incorporating mechanical properties into the VAE did not have a significant effect on the performance of geometric smoothness, which partially describes the ability to produce connected transition regions. This suggests that incorporating mechanical properties into a data-driven model does not affect the connectivity of adjacent unit cells in a transition region, design criterion ①. Alternatively, the ability of the model to address mechanical properties of adjacent unit cells, design criterion ③, was affected by incorporating mechanical properties into the embedding of the model. The effects of the latent space distance on stiffness continuity were reduced, which is desirable in order to offer more control over the performance of stiffness continuity. The results from the hybrid VAE indicate that its latent space is better suited to create multi-lattice transition with continuous stiffness properties, design criterion ③, given that latent space distance has a reduced impact on its performance.

## 4. Conclusion

The development of multi-lattice transition regions is increasingly dominated by generative machine learning models. These models aim to produce multi-lattice transition regions with smooth geometric and mechanical properties by identifying ideal sets of unit cells to use in density-based topology optimization. This work compares two approaches to choosing these sets of unit cells: a model that uses only geometric information and another hybrid model that merges geometry and property information. The first model, the geometry VAE, created a geometrically defined latent space only using the unit cell topologies. The second model, the hybrid VAE, utilized a combination of unit cell topologies and unit cell stiffnesses to define the latent space. These models are evaluated on their ability to address two design criteria: ① maintain connectivity between adjacent unit cells[20] and ③ design based on the mechanical characteristics of adjacent unit cells.[27]

In summary, our analysis indicates that both the geometry-based and hybrid Variational Autoencoders (VAEs) exhibited similarities in their performance. It was again observed that geometric smoothness is negatively impacted by latent space distance, while being positively influenced by the interaction between latent space distance and transition length. Interestingly, the incorporation of mechanical properties into the hybrid VAE did not significantly alter its ability to develop geometrically smooth transition regions, which is partially related to design criterion ①. Alternatively, the models performed differently regarding stiffness continuity through the transition region. The hybrid VAE outperformed the geometry-based VAE, showing that stiffness continuity was less adversely affected by latent space distance in the hybrid embedding. This suggests that the hybrid VAE is better suited for design tasks that require the maintenance of smooth mechanical properties through the transition region, aligning with design criterion ③. Nevertheless, it's worth noting that approximately 10% of the variability in stiffness continuity remains unaccounted for, which suggests the existence of additional variables that further describe the performance of stiffness continuity. Overall, the major contribution of this work is increasing the ability of a VAE to generate structures with smoothly changing physical properties, per design criterion ③.

The current work focused on evaluating a set of 2-dimensional unit cells but future work should explore 3-dimensional unit cells, manufacturability evaluations, and methods for applying these unit cells to a structure. Although the applicability of 2-dimensional unit cells is limited, this work identified the clustering effects as a potential issue before transitioning into 3-dimensional unit cells. The clustering is likely caused by diversity in the dataset and needs to be addressed in order to create useful sets of unit cells. Future work will explore methods to address the effects of clustering while comparing performance based on the metrics established for geometry and stiffness properties.[33,34] Throughout the work, density-based topology optimization has been considered the main approach to applying a set of unit cells to a multi-lattice structure.[13,20,25,26,28] However, it is unclear whether this is the best approach to creating multi-lattice structures, which indicates that further research into the creation of multi-lattice structures is necessary. Finally, there are many ways in which property information can be combined with geometric information in the training process of hybrid models. Therefore, future work should also explore the design freedom that is present in the design space of deep learning models.

**Authorship Contribution Statement:**
Martha Baldwin: Conceptualization (equal), Data Curation (lead), Formal Analysis (equal), Investigation (lead), Methodology (lead), Project Administration (equal), Software (lead), Validation (lead), Visualization (lead), Writing – original draft (lead)

Dr. Chris McComb: Conceptualization (equal), Formal Analysis (equal), Methodology (supporting), Project Administration (equal), Resources (lead), Supervision (lead), Writing – review & editing (lead)

Dr. Nicholas Meisel: Conceptualization (equal), Supervision (supporting), Writing – review & editing (supporting)

**Author Disclosure Statement:**
No competing financial interests exist.

**Funding Statement:**
This material is based upon work supported by the National Science Foundation through Grant No. CMMI-1825535. Any opinions, findings, conclusions, or recommendations expressed in this paper are those of the authors and do not necessarily reflect the views of the sponsors.

**Acknowledgements:**
We are grateful to Yigitcan Comlek from the Integrated DEsign Automation Laboratory (IDEAL) at Northwestern University for generously sharing their data to support this work.

# References

1. Hanks B, Berthel J, Frecker M, et al. Mechanical properties of additively manufactured metal lattice structures: Data review and design interface. Addit Manuf 2020;35; doi: 10.1016/J.ADDMA.2020.101301.

2. Plocher J, Panesar A. Effect of density and unit cell size grading on the stiffness and energy absorption of short fibre-reinforced functionally graded lattice structures. Addit Manuf 2020;33:101171; doi: 10.1016/J.ADDMA.2020.101171.

3. Kantareddy SNR, Roh BM, Simpson TW, et al. Saving Weight with Metallic Lattice Structures: Design Challenges with Real-World Example. In: Paper Presented at 27th Annual International Solid Freeform Fabrication Symposium - An Additive Manufacturing Conference 2016.

4. Panesar A, Abdi M, Hickman D, et al. Strategies for functionally graded lattice structures derived using topology optimisation for Additive Manufacturing. Addit Manuf 2018;19:81–94; doi: 10.1016/J.ADDMA.2017.11.008.

5. Rossiter JD, Johnson AA, Bingham GA. Assessing the Design and Compressive Performance of Material Extruded Lattice Structures. 3D Print Addit Manuf 2020;7; doi: 10.1089/3dp.2019.0030.

6. Lozanovski B, Downing D, Tino R, et al. Image-Based Geometrical Characterization of Nodes in Additively Manufactured Lattice Structures. 3D Print Addit Manuf 2021;8; doi: 10.1089/3dp.2020.0091.

7. Niknam H, Akbarzadeh AH. Graded lattice structures: Simultaneous enhancement in stiffness and energy absorption. 2020; doi: 10.1016/j.matdes.2020.109129.

8. Li D, Dai N, Tang Y, et al. Design and Optimization of Graded Cellular Structures with Triply Periodic Level Surface-Based Topological Shapes. Journal of Mechanical Design 2019;141(7); doi: 10.1115/1.4042617/727182.

9. Wang Y, Zhang L, Daynes S, et al. Design of graded lattice structure with optimized mesostructures for additive manufacturing. Mater Des 2018;142:114–123; doi: 10.1016/J.MATDES.2018.01.011.

10. Goel A, Anand S. Design of Functionally Graded Lattice Structures using B-splines for Additive Manufacturing. Procedia Manuf 2019;34:655–665; doi: 10.1016/J.PROMFG.2019.06.193.

11. Maskery I, Hussey A, Panesar A, et al. An investigation into reinforced and functionally graded lattice structures: Journal of Cellular Plastics 2016;53(2):151–165; doi: 10.1177/0021955X16639035.